\documentclass{article} 
\usepackage{amssymb}
\usepackage{amsmath}
\usepackage{amsthm}
\usepackage[a4paper]{geometry}
\usepackage[round]{natbib}
 \usepackage[colorlinks=true, linkcolor=blue]{hyperref}

\usepackage{amssymb}
\usepackage{amsmath,amscd}
\usepackage{amsthm}
\usepackage{cancel}     \usepackage{centernot}  \usepackage{graphicx}   \usepackage{soul}       \usepackage{color}      \usepackage[shortlabels, inline]{enumitem}   \usepackage{commath}    \usepackage{subcaption} \usepackage{booktabs}   
\usepackage{pgfplots}
\usepackage{tikz}
\usetikzlibrary{positioning, arrows.meta, fit, backgrounds}
\usepackage[ruled, vlined]{algorithm2e}

\SetKw{Continue}{continue}
\SetKw{Break}{break}
\usepackage{float}
\newfloat{program}{thp}{lop}
\floatname{program}{Algorithm}
\newtheorem{thm}{Theorem}[section]
\newtheorem{lemma}[thm]{Lemma}
\newtheorem{cor}[thm]{Corollary}
\newtheorem{prop}[thm]{Proposition}

\newtheorem{fact}{Fact}[section]
\newtheorem*{thm*}{Theorem}
\newtheorem*{lemma*}{Lemma}
\newtheorem*{cor*}{Corollary}
\newtheorem*{prop*}{Proposition}
\newtheorem*{conjecture*}{Conjecture}
\theoremstyle{definition}
\newtheorem{defn}{Definition}[section]
\newtheorem*{defn*}{Definition}
\theoremstyle{definition}

\theoremstyle{definition}

\theoremstyle{remark}
\newtheorem*{ex*}{Example}
\theoremstyle{definition}

\theoremstyle{definition}
\newtheorem*{assm*}{Assumption}
\theoremstyle{remark}
\newtheorem{remark}{Remark}[section]
\theoremstyle{remark}
\newtheorem*{remark*}{Remark}
\usepackage{xcolor}
\DeclareFontFamily{U}{mathx}{\hyphenchar\font45}
\DeclareFontShape{U}{mathx}{m}{n}{
      <5> <6> <7> <8> <9> <10> gen * mathx
      <10.95> mathx10 <12> <14.4> <17.28> <20.74> <24.88> mathx12
      }{}
\DeclareSymbolFont{mathx}{U}{mathx}{m}{n}
\DeclareMathSymbol{\intop}  {1}{mathx}{"B3}
\DeclareFontFamily{U}{mathx}{\hyphenchar\font45}
\DeclareFontShape{U}{mathx}{m}{n}{
      <5> <6> <7> <8> <9> <10>
      <10.95> <12> <14.4> <17.28> <20.74> <24.88>
      mathx10
      }{}
\DeclareSymbolFont{mathx}{U}{mathx}{m}{n}
\DeclareFontSubstitution{U}{mathx}{m}{n}
\DeclareMathAccent{\widecheck}{0}{mathx}{"71}
\DeclareMathAccent{\wideparen}{0}{mathx}{"75}

\newcommand\indep{\independent}
\newcommand\independent{\protect\mathpalette{\protect\independenT}{\perp}}
\def\independenT#1#2{\mathrel{\rlap{$#1#2$}\mkern4mu{#1#2}}}

\newcommand{\wh}{\widehat}
\newcommand{\ol}{\overline}

\let\temp\phi
\let\phi\varphi
\let\varphi\temp

\renewcommand{\sec}{\textsection}

\newcommand{\pr}{\mathbb{P}}

\newcommand{\R}{\mathbb{R}}

\newcommand{\E}{\mathbb{E}}

            \newcommand{\given}{\,|\,}

\newcommand{\eps}{\varepsilon}

\newcommand{\iid}{\overset{\text{iid}}{\sim}}

\DeclareMathOperator*{\argmax}{arg\,max}
\DeclareMathOperator{\dist}{dist}

\DeclareMathOperator{\HH}{H}

\newcommand{\probm}{\rho}

\newcommand{\dH}{\probm_{\HH}}

\newcommand{\gr}{\mathsf{G}}
\DeclareMathOperator{\pa}{pa}   \DeclareMathOperator{\nd}{nd}

\newcommand{\AS}{\text{ as }}
\newcommand{\WHERE}{\text{ where }}

\newcommand{\FORALL}{\text{ for all }}

\usepackage{aligned-overset}

\newcommand{\I}{[0,1]}
\newcommand{\ind}{\mathcal{I}}

\DeclareMathOperator{\BF}{BF}
\DeclareMathOperator{\odds}{PR}

\newcommand{\fcns}{\mathcal{F}}

\DeclareMathOperator{\nb}{nb}

\newcommand{\fcn}{f}
\newcommand{\fcng}{g}
\newcommand{\cfcn}{\ol{\fcn}}
\newcommand{\cfcng}{\ol{\fcng}}
\newcommand{\eq}{\mathsf{C}}
\newcommand{\grH}{\mathsf{H}}
\newcommand{\ver}{\mathsf{V}}
\newcommand{\edg}{\mathsf{E}}
\renewcommand{\sp}{s}
\newcommand{\test}{\phi}

\newcommand{\level}{\lambda}
\DeclareMathOperator{\dags}{\textup{\textsf{DAG}}}
\DeclareMathOperator{\DAG}{\textup{\textsf{DAG}}}

\DeclareMathOperator{\BIC}{BIC}

\newcommand{\model}{\mathcal{M}}

\newcommand{\dens}{\mathcal{D}}
\newcommand{\fcnm}{\overline{\probm}}

\newcommand{\revidx}{t_{0}}
\newcommand{\lastidx}{r}
\newcommand{\gesout}{\wh{\gr}}
\renewcommand{\dist}{P}
\newcommand{\distn}{P^{n}}
\newcommand{\distf}{\dist_{\fcn}}
\newcommand{\jointd}{p}
\newcommand{\jointdf}{\jointd_{\fcn}}
\newcommand{\true}{\dist^{*}}
\newcommand{\truen}{\dist^{*,n}}
\newcommand{\trued}{\jointd^{*}}
\newcommand{\truegr}{\gr^{*}}

\newcommand{\truefcn}{\fcn^{*}}
\newcommand{\truef}{\dist_{\truefcn}}
\newcommand{\truedf}{\jointd_{\truefcn}}
\newcommand{\jdf}{\jointdf}

\newcommand{\tg}{\truegr}

\newcommand{\tf}{\truefcn}

\newcommand{\indepx}[1]{\indep_{\!\!#1}\,}

\newcommand{\indepP}{\indep_{\!\!\dist}\,}

\newcommand{\prior}{\pi}
\renewcommand{\pr}{\prior} 
\newcommand{\reg}{\gamma}
\newcommand{\lip}{L}

\newcommand{\lipclass}{\mathcal{H}^{1,\lip}}

\newcommand{\fcnsl}{\fcns^{1,\lip}}
\newcommand{\nullgr}{\gr_{\emptyset}}

\newcommand{\me}{\eps}

\newcommand{\prconst}{\Gamma}
\newcommand{\prconstj}{\gamma}
 
\title{Greedy equivalence search for nonparametric graphical models}
\author{Bryon Aragam}
\date{\emph{University of Chicago}}
\begin{document}
\maketitle

{\let\thefootnote\relax\footnote{Contact: \texttt{bryon@chicagobooth.edu}}}

\begin{abstract}
One of the hallmark achievements of the theory of graphical models and Bayesian model selection is the celebrated greedy equivalence search (GES) algorithm due to Chickering and Meek. GES is known to consistently estimate the structure of directed acyclic graph (DAG) models in various special cases including Gaussian and discrete models, which are in particular curved exponential families. A general theory that covers general nonparametric DAG models, however, is missing. Here, we establish the consistency of greedy equivalence search for general families of DAG models that satisfy smoothness conditions on the Markov factorization, and hence may not be curved exponential families, or even parametric. The proof leverages recent advances in nonparametric Bayes to construct a test for comparing misspecified DAG models that avoids arguments based on the Laplace approximation. Nonetheless, when the Laplace approximation is valid and a consistent scoring function exists, we recover the classical result. As a result, we obtain a general consistency theorem for GES applied to general DAG models.
\end{abstract}

\section{Introduction}
\label{sec:intro}

Bayesian networks are graphical models specified by directed acyclic graphs (DAGs), and can be used to represent distributions, model causal relationships, enable probabilistic reasoning, and encode information flow through a network.
Each DAG defines a canonical statistical model, i.e. a collection of multivariate probability distributions satisfying the Markov property implied by the DAG. 
Often we do not know the structure of this model, which is crucial in the aforementioned applications.
Instead, all we have are
samples from a distribution, and we first need to learn the structure of the DAG that defines the model. 
This is the classical \emph{structure learning} problem that has attracted significant attention over the past several decades. This problem subsumes and generalizes several well-known model selection problems such as variable selection and conditional independence testing. Our interest here is in nonparametric aspects of structure learning.

Broadly speaking, existing approaches to structure learning fall into two categories (along with hybrid approaches that combine aspects of both): Constraint-based methods, based on conditional independence testing, and score-based methods, based on global optimization. 
For constraint-based methods, the well-known PC algorithm \citep{spirtes1991} is consistent for general, nonparametric models as long as a consistent conditional independence test is available. Indeed, nonparametric and high-dimensional variants of the PC algorithm have been around for some time \citep[e.g.][]{kalisch2007,harris2013,harris2012}.
On nonparametric conditional independence testing in particular, there is a rapidly developing literature \citep[see e.g.][and the references therein]{shah2018hardness,canonne2018testing,azadkia2019simple,neykov2020minimax}.

By contrast, for score-based as well as closely related Bayesian approaches to structure learning, less is known in nonparametric settings.
Perhaps the most well-known score-based algorithm is greedy equivalence search (GES), which was proposed by \citet{meek1997thesis} and proved consistent by \citet{chickering2002ges} when the graphical models under consideration are curved exponential families. The proof relies on the consistency of the Bayesian information criterion (BIC) in these families, see e.g. \citet{haughton1988}.
More recently, \citet{nandy2018} proved a notable high-dimensional consistency result for GES for linear models.

Once the models under consideration become nonparametric, however, the consistency of a GES-like greedy search algorithm is less clear.
We show that a suitable modification of GES is consistent for selecting nonparametric directed acyclic graphical models under weak smoothness (i.e. Lipschitz or H\"older) conditions on the Markov factorization. In particular, we do not involve additivity constraints, independent and/or Gaussian noise, linearity, or other semiparametric assumptions that have become common in the structure learning literature. Moreover, our proposed modification retains the greedy nature of the original GES algorithm, and only requires models in a neighbourhood (defined by single edge removals and additions) to be compared in each step.

To clarify our statistical setting, we outline the basic model here, while deferring precise details to Section~\ref{sec:bg}.
Given a random vector $X=(X_{1},\ldots,X_{d})$ and its joint distribution $\dist(X)$, we assume $\dist$ has a density $\jointd$ with respect to some base measure. The distribution $\dist$ is assumed to be Markov to some directed acyclic graph (DAG) $\gr=(\ver,\edg)$, i.e. 
the joint density $\jointd(x)$ factors as the product of $d$ local conditional probability densities (CPDs) $\fcn=(\fcn_{1},\ldots,\fcn_{d})$:
\begin{align}
\label{eq:joint:model}
\jointd(x)
= \prod_{j=1}^{d} \fcn_{j}(x_{j}\given\pa(j)),
\quad
\pa(j)
:= \{k : k\to j\}
=\{ k\in\ver: (k,j)\in\edg\}.
\end{align}
Here, we are as usual abusing notation by identifying the vertex set with $X$ or equivalently the index set $[d]:=\{1,\ldots,d\}$, i.e. $\ver=X=[d]$. The choice to write $\fcn_{j}(x_{j}\given\pa(j))$ instead of the usual $\jointd(x_{j}\given\pa(j))$ is intentional; the reasons for this will become clear in the sequel. 
The family of densities that satisfy \eqref{eq:joint:model} defines a graphical model, denoted by $\model(\gr)$.
When we wish to emphasize the dependence of $\dist$ (resp. $\jointd$) on $\fcn$, we will write $\dist=\distf$ (resp. $\jointd_{\fcn}$), although this dependence will often be suppressed. Observe that the map $\fcn\mapsto \distf$ is not one-to-one, which is an issue that GES and related structure learning algorithms must contend with.

Now suppose a distribution $\true$ satisfies \eqref{eq:joint:model} for some DAG $\truegr$ with factorization $\tf=(\tf_{1},\ldots,\tf_{d})$, i.e. $\true=\truef\in\model(\truegr)$.
Given $n$ i.i.d. observations $X^{(i)}\iid \true$, the problem of interest is to learn the structure of the DAG $\truegr$ up to Markov equivalence. 
Equivalently, we wish to identify and select the directed graphical model $\model(\truegr)$ from data.
As is well-known, even ensuring this problem is well-defined is complicated, owing to Markov equivalence and the non-identifiability of $\tg$ (as well as $\tf$ noted above) from $\true$, unless further assumptions are made.
In the special case where each conditional distribution $\tf_{j}$ can be expressed as a linear model, we obtain a linear structural equation model, which has been extensively studied in the literature. As mentioned above, it is known that GES is consistent for linear Gaussian models \citep{chickering2002ges,nandy2018}.  Our main interest will be the case where the $\tf_{j}$ are general conditional densities (e.g. there may not even be additive errors), and hence these existing results no longer apply.

Unfortunately, without \emph{any} assumptions on the model, nonparametric model selection is fundamentally difficult. 
This has been formalized in the setting of conditional independence testing \citep{shah2018hardness,neykov2020minimax}.
Our main result is that Lipschitz continuity suffices to consistently estimate $\tg$.
Crucially, we do not assume that the conditional densities $\tf_{j}$ have any particular form.
We also do not prescribe assumptions on the noise distribution (e.g. Gaussian) or structure (e.g. independence, additivity, etc.). To help keep things concrete, we mention here that our assumptions subsume most commonly used models in the structure learning literature, and are broadly satisfied by general families of models, including spline densities, infinite-dimensional exponential families, smooth nonparanormal models, causal additive models, additive noise models, and post-nonlinear models.

\subsection{Related work}
\label{sec:relatedwork}

Early work establishing the foundations for learning Bayesian networks includes \citep{rebane1987recovery,herskovits1990kutato,buntine1991theory,spirtes1991,cooper1992bayesian,bouckaert1993probabilistic,spiegelhalter1993bayesian,heckerman1995learning}. 
Since a comprehensive review of the history of this problem is beyond the scope, we focus our review on the most closely related work below.
For a detailed review of the history and methods in Bayesian network structure learning, we refer the reader to textbooks such as \citet{spirtes2000,neapolitan2004learning,peters2017elements} and review articles by \citet{drton2016} and \citet{squires2023causal}. See also \citet{heckerman2008tutorial}.

Greedy algorithms for model search of course abound in the literature, and include classical methods for subset selection such as forward and backward stepwise regression. A review of this literature can be found in \citet{miller2002subset}, and more recently, \citet{bertsimas2016best}. Greedy strategies for Bayesian model selection in graphical models appeared as early as the work of \citet{raftery1993bayesian,madigan1994model}. Later, in his PhD thesis, \citet{meek1997thesis} introduced GES as a formal procedure for model search. The consistency of GES for Gaussian and multinomial models was proved by \citet{chickering2002ges}.
The greedy approach taken by GES should be contrasted with Bayesian approaches that estimate a full posterior on the space of DAG models \citep{madigan1994model,madigan1995bayesian,madigan1996bayesian,giudici2003improving,friedman2003,koivisto2004exact,ellis2008,kuipers2017partition,agrawal2018minimal}.

Several notable developments have occurred since this pioneering work on GES.
Although it is consistent, the worst-case computational and statistical complexity of GES leaves much to be desired (despite, somewhat surprisingly, being quite fast and accurate on practical problems). To address this, \citet{chickering2015selective} introduced \emph{selective} GES to improve the computational complexity by reducing the number of score evaluations; later \citet{chickering2020statistically} introduced \emph{statistically efficient} GES to reduce the size of the conditioning sets used by GES. 
\citet{ramsey2016} showed that GES can scale to problems with millions of variables, with further computational refinements in \citet{nazaret2024extremely}.
\citet{nandy2018} proved the high-dimensional consistency of GES under bounds on the growth of so-called ``oracle versions'', and
\citet{zhou2023complexity} analyzed the complexity of MCMC-based algorithms for Bayesian learning of Gaussian DAG models by leveraging aspects of GES as a greedy search procedure. 
\citet{solus2017} consider geometric aspects of greedy search, and propose a greedy permutation-based algorithm for recovering the true Markov equivalence class. Their results presume the usual regularity conditions needed for CI testing and the BIC score. Further generalizations can be found in \citet{lam2022greedy} and \citet{linusson2023greedy}.
While not explicitly about GES, there has also been related work on high-dimensional score-based algorithms such as the MLE \citep{geer2013} and least-squares loss \citep{loh2014causal,aragam2019globally}. 
\citet{evans2018model,guo2020testing} show how model selection between parametric DAG models suffers from statistical deficiencies not shared by other model selection problems, and in particular the geometry of the model selection problem limits the power of any test for distinguishing parametric DAG models.

In contrast to this existing work, our focus is on understanding nonparametric aspects, under which classical regularity conditions fail.
More recently, there has been work along these lines more closely related to the results presented here. 
Causal additive models \citep[CAM,][]{buhlmann2014} are a notable nonparametric model, however, the greedy search heuristic for CAM lacks formal guarantees. By contrast, our version of GES has formal consistency guarantees and applies to more general models without additive structure or Gaussian noise.
\citet{huang2018generalized} introduced a decomposable and locally consistent scoring criterion RKHS embeddings; their results crucially exploit Gaussianity and additivity in the noise in order to define a score. 
\citet{shen2022reframed} propose a neural conditional dependence measure and show how to integrate this into GES, and establish its consistency under parametric (i.e. finite-dimensional) assumptions. More recent developments include \citet{schauer2023causal,caron2023structure}.

Finally, although DAGs are known to present unusual challenges in structure learning, other types of graphical models have a well-developed theory for nonparametric structure learning. For example, in undirected graphs there is the nonparanormal model \citep{liu2009,lafferty2012,liu2012high,xue2012regularized} and subgaussian inverse covariance models \citep{ravikumar2011}.
Also worth mentioning are extensions to exponential family graphical models \citep{yang2015}. Compared to the present work, there are two key distinctions: 1) Our focus is on DAG models, for which the analysis is significantly more involved (e.g. the graph cannot be derived from covariance or other second-order statistics), and 2) Our assumptions, which involve only smoothness conditions, are more general than e.g. nonparanormal.

\subsection{Overview}

Our main contribution is to show that (a suitably modified) GES algorithm consistently recovers the Markov equivalence class of nonparametric DAG models as defined in \eqref{eq:joint:model}; see Theorem~\ref{thm:ges:cons} for a precise statement. 
More generally, in the absence of faithfulness conditions, GES still recovers a minimal model.
We make no assumptions on the form of conditional densities or the Markov factorization, beyond smoothness assumptions. 
At the same time, we show that some assumptions are necessary by extending existing no-free-lunch theorems to graphical models (Section~\ref{sec:gap}).

The main idea is to replace the BIC score, which may no longer be well-defined for general infinite-dimensional models, with the posterior odds ratio (closely related to the commonly used Bayes factor), and instead of scoring individual models, using a Bayesian model selection procedure to compare models (Section~\ref{sec:bayes}). 
This allows us to consider general nonparametric families under standard smoothness conditions. Moreover, when the model is sufficiently regular (i.e. so as to admit a BIC-style approximation to the Bayes factor), we recover the classical result for parametric models. Thus, our assumptions cover the majority of parametric and semiparametric models that appear in the literature on structure learning, including Gaussian models \citep{kalisch2007,nandy2018}, exponential families \citep{chickering2002ges,haughton1988}, additive noise models \citep{peters2014}, post-nonlinear models \citep{zhang2009}, linear non-Gaussian models \citep{shimizu2006}, 
and smooth nonparanormal models \citep{harris2013}.

\section{Preliminaries}
\label{sec:bg}

In this section we describe some preliminary notation and background to make these ideas precise. For a detailed review of this material, we refer to the reader to \citet{lauritzen1996,koller2009}.

\paragraph{Notation}
Sets of vertices, indices, and random variables will be treated the same, so $A\subset[d]$ can be viewed as a set of indices, a set of vertices, or a set of random variables in the obvious way.
For any $A\subset[d]$, $X_{A}:=\{X_{j}:j\in A\}$.
Set union will often be denoted by juxtaposition, i.e. $AB:=A\cup B$, and this notation extends to singletons, e.g. $123=\{1,2,3\}$.
$1(A)$ is the indicator function of the set $A$. Conditional independence in the distribution $\dist$ is denoted by $\indepx{\dist}$, and $d$-separation in the DAG $\gr$ is denoted by $\indepx{\gr}$. The $n$-fold product measure of a distribution $\dist$ is denoted by $\distn$; we also use the shorthand $\truen:=(\true)^{n}$.
We also make use of standard multi-index notation for denoting partial derivatives: We let $\alpha=(\alpha_{1},\ldots,\alpha_{d})$ be a multi-index and $D^{\alpha}f$ denotes the $\alpha$-mixed partial derivative of $f$, and $|\alpha|:=\alpha_{1}+\cdots+\alpha_{d}$.
Since $\fcn_{j}:\R^{|\pa(j)|+1}\to\R$ is a function of $|\pa(j)|+1$ variables, it can and will be written $\fcn_{j}(x_{j},\pa(j))$.
To avoid overloading notation, we will often write $\fcn_{j}(x)$ instead of $\fcn_{j}(x_{j}\given\pa(j))$ or $\fcn_{j}(x_{j},\pa(j))$, and use these interchangeably.

\subsection{Graphical models}
\label{sec:bg:gm}
We use standard terminology and notation from the literature on graphical models.
The set of all DAGs over $\ver=X$ will be denoted by $\DAG$.
A DAG $\gr=(\ver,\edg)$ is a directed graph that does not contain any directed cycles.
When we wish to emphasize the vertex and/or edge set corresponding to $\gr$, we will write $\ver(\gr)$ and/or $\edg(\gr)$, respectively. Edges may be denoted by $(k,j)\in\edg$ or $k\to j\in\edg$, and where no confusion may arise, we will also write $(k,j)\in\gr$ or $k\to j\in\gr$.
We write $\gr\subset \grH$ if and only if $\ver(\gr)=\ver(\grH)$ and $\edg(\gr)\subset \edg(\grH)$.

If $X \rightarrow Y$ is an edge, then $Y$ is called the $\emph{child}$ of $X$; the node $X$ is called the \emph{parent} of $Y$, and $X$ and $Y$ are said to be \emph{adjacent} to each other. 
The set of parents of a node $X$ is denoted by $\pa(X)$. 
We say that $Y$ is a \emph{descendant} of $X$, and $X$ is an \emph{ancestor} of $Y$ if there exists a directed path from $X$ to $Y$. 
If $Y$ is a not a descendant of $X$ (i.e. there is no directed path from $X$ to $Y$), we call $Y$ a \emph{non-descendant} of $X$ and denote the set of non-descendants of $X$ by $\nd(X)$. 
A graph is \emph{complete} if for any pair of nodes $X, Y$, we have either $X \rightarrow Y$ or $Y \rightarrow X$. A subgraph is called a \emph{v-structure} if it is of the form $X \rightarrow Y \leftarrow Z$, and we call the child node $Y$ a \emph{collider}. 
We say that a collier is \emph{unshielded} if the parents of the collider do not share an edge.  
Given a directed graph $\gr$, we call the undirected graph that is obtained by replacing each directed edge with an undirected edge the \emph{skeleton} of $\gr$.

A minimal I-map of $\dist$ is any DAG $\gr=(\ver,\edg)$ such that the following conditions hold:
\begin{enumerate}
\item $\dist$ factorizes over $\gr$, i.e. \eqref{eq:joint:model} holds, and
\item If any edge is removed from $\edg$, then \eqref{eq:joint:model} is violated.
\end{enumerate}
A \emph{trail} in a directed graph $\gr$ is any sequence of distinct nodes $X_{i_{1}},\ldots,X_{i_{\ell}}$ such that there is an edge between $X_{i_{m}}$ and $X_{i_{m+1}}$. The orientation of the edges does not matter. For example, $X\leftarrow Y\rightarrow Z$ would be a valid trail. 
A trail of the form $X_{i_{m-1}}\rightarrow X_{i_{m}}\leftarrow X_{i_{m+1}}$ is called a \emph{$v$-structure}.
We call a trail $t$ \emph{active} given another set $C$ if (a) $X_{i_{m}}\in C$ for every $v$-structure $X_{i_{m-1}}\rightarrow X_{i_{m}}\leftarrow X_{i_{m+1}}$ in $t$ and (b) no other node in $t$ is in $C$. In other words, $t\cap C$ consists only of colliders in some $v$-structure contained entirely in $t$.
\begin{defn}
Let $A,B,C$ be three sets of nodes in $\gr$. We say that $A$ and $B$ are $d$-separated by $C$ if there is no active trail between any node $a\in A$ and $b\in B$ given $C$.
\end{defn}
\noindent
An important consequence of \eqref{eq:joint:model} is the following: If $\gr$ satisfies \eqref{eq:joint:model} for some $\dist$ and $A$ and $B$ are $d$-separated by $C$ in $\gr$, then $A\indepP B\given C$ (\citealp{koller2009}, Theorems~3.1, 3.2, or \sec3.2.2 in \citealp{lauritzen1996}).
Thus, if $\gr$ is a BN of $\dist$, then $d$-separation can be used to read off a subset of the conditional independence relations in $\dist$. 
When $A$ is $d$-separated
from $B$ given $C$ in the DAG $\gr$, we write $A\indepx{\gr}B\given C$.

A DAG $\gr\in\DAG$ defines two objects of interest: A set of $d$-separation statements, denoted by $\ind(\gr)$, and a statistical model, denoted by $\model(\gr)$. For the former, we have 
\begin{align}
\label{eq:def:Imodel}
\ind(\gr)
:= \{ (A,B,C)\subset\ver\times\ver\times\ver : A\indepx{\gr}B\given C \}.
\end{align}
A distribution $\dist$ also defines a set of conditional independence (CI) relationships, i.e.
\begin{align}
\label{eq:def:Imodel}
\ind(\dist)
:= \{ (A,B,C)\subset\ver\times\ver\times\ver : A\indepP B\given C\}.
\end{align}
We say that $\dist$ is Markov to $\gr$ if $\ind(\gr)\subset\ind(\dist)$ and faithful to $\gr$ if $\ind(\dist)\subset\ind(\gr)$; when both inclusions hold $\gr$ is called a perfect map of $\dist$.
This defines the statistical model $\model(\gr)$ as the set of all joint densities $\jointd$ whose law $\dist$ is Markov to $\gr$, i.e.
\begin{align}
\label{eq:def:Gmodel:1}
\model(\gr)
&= \{\jointd : \dist\text{ is Markov wrt to $\gr$} \} \\
\label{eq:def:Gmodel:2}
&= \{\jointd : \ind(\gr)\subset\ind(\dist) \} \\
\label{eq:def:Gmodel:3}
&= \{\jointd : \text{factorization \eqref{eq:joint:model} holds}\}.
\end{align} 
We will often conflate $\gr$ with $\model(\gr)$, referring to $\gr$ as a model whenever there is no confusion.

Recall the definition of the parent set $\pa(j)$ in \eqref{eq:joint:model}. 
A DAG is uniquely specified by its parent sets, which we can write inline as follows: $$\gr = [\pa_{\gr}(1)\given\pa_{\gr}(2)\given\cdots\given\pa_{\gr}(d)].$$ For example, a Markov chain on 3 nodes is given by $\gr=[\emptyset\given1\given2]$ and a complete DAG by $\gr=[\emptyset\given1\given12]$.
The \emph{sparsity sequence} of $\gr$ is the sequence $\sp=\sp(\gr)=(\sp_{1},\ldots,\sp_{d})$ with $\sp_{j}:=|\pa_{\gr}(j)|$. 
For any $(k,j)\in\ver\times\ver$, we define:
\begin{itemize}
\item $\gr_{kj}$ to be the DAG that results from adding the edge $k\to j$ to $\gr$; if $k\to j\in\gr$ already, then $\gr_{kj}=\gr$.
\item $\gr_{-kj}$ to be the DAG that results from removing the edge $k\to j$ from $\gr$; if $k\to j\notin\gr$ already, then $\gr_{-kj}=\gr$.
\end{itemize}

It is a standard fact that a distribution may belong to more than one model $\model(\gr)$, and the equivalence relation induced by model equivalence is known as Markov equivalence. There is a well-known graphical characterization of model equivalence in terms of the skeleton and colliders in a DAG:
\begin{fact}
\label{fact:equiv}
$\model(\gr)=\model(\gr')$ if and only if $\gr$ and $\gr'$ share the same skeleton and unshielded colliders.
\end{fact}
\noindent
Markov equivalence defines an equivalence relation, which gives rise to the CPDAG (completed partially directed acyclic graph) representation of the MEC. The CPDAG is simply a partial DAG (PDAG) such that $k\to j$ if and only if $k\to j$ in every DAG in the equivalence class, and $k-j$ otherwise.
When $\dist$ admits a perfect map, all of the CI relations in $\dist$ can be read off via $d$-separation in its underlying CPDAG and the CPDAG becomes the estimand of interest in structure learning. 

With this in mind, to any DAG $\gr$ we may associate its corresponding CPDAG $\eq(\gr)$; thus $\model(\gr)=\model(\gr')$ if and only if $\eq(\gr)=\eq(\gr')$. With some abuse of notation, we write $\grH\in\eq(\gr)$ to indicate that $\eq(\grH)=\eq(\gr)$. A key component of GES is its local search space, which are the neighbourhoods of a DAG $\gr$. Instead of considering single-edge additions (or removals) to $\gr$, GES considers single-edge additions to the MEC of $\gr$, i.e. $\eq(\gr)$. Formally, we define the following neighbourhoods of $\gr$:
\begin{align}
\label{eq:def:nbplus}
\nb_{+}(\gr)
&:=\bigcup_{\grH\in\eq(\gr)}\{\grH_{kj} : (k,j)\notin\edg(\grH)\}, \\
\label{eq:def:nbminus}
\nb_{-}(\gr)
&:=\bigcup_{\grH\in\eq(\gr)}\{\grH_{-kj} : (k,j)\notin\edg(\grH)\}
\end{align}
In others words, a DAG $\gr'$ is in $\nb_{+}(\gr)$ (resp. $\nb_{-}(\gr)$) if and only if there is some DAG $\grH\in\eq(\gr)$ to which we can add (resp. remove) a single edge that results in $\gr'$. See \citet{chickering2002ges} for details, including efficient algorithms for the computation of $\nb_{\pm}(\gr)$.

\subsection{Function spaces and smoothness}
\label{sec:bg:fcn}

In parametric graphical models, we attach to each DAG $\gr$ a parameter $\theta\in\R^{p}$ that fully parametrizes a distribution $\dist_{\theta}$ that is Markov to $\gr$. For example, in a Gaussian structural equation model $\theta$ corresponds to the coefficients $\beta_{kj}$ (or each edge $k\to j\in\edg$) and conditional variances $\sigma_{j}^{2}$ (for each node $j\in\ver$) for a particular distribution. In the nonparametric setting, there is no longer a finite-dimensional parameter $\theta\in\R^{p}$, and instead the model is parametrized directly by the local conditional probability densities (CPDs) themselves. Recall that these CPDs are encapsulated by the parameter $\fcn=(\fcn_{1},\ldots,\fcn_{d})$. 
In order to establish control over local neighbourhoods during the search phase, instead of working with joint densities in $\model(\gr)$, we will work on the space of CPD factorizations $\fcns(\gr)$ defined as follows.

First, to simplify the presentation and keep our focus on the novel, graphical aspects of this problem,
we assume for now that all densities are Lipschitz continuous and supported on $\I^{d}$. See also Remark~\ref{rem:assm}.
We do not impose any other functional or parametric restrictions on the form of $\fcn$.
For $\lip\ge0$, we let $\lipclass(\I^{m})$ denote the space of $\lip$-Lipschitz functions $\fcn:\I^{m}\to\R$. 
Now define (we often suppress the dependence on $\lip$ to reduce notational burden)
\begin{align*}
\fcns_{j}(\gr)
= \fcnsl_{j}(\gr)
:= \bigg\{\fcn_{j}\in\lipclass(\I^{\sp_{j}+1}) : \fcn_{j} > 0,\, \int \fcn_{j}(x_{j},x_{\pa_{\gr}(j)})\dif{x_{j}}=1 \FORALL x_{\pa_{\gr}(j)}\in\I^{\sp_{j}} \bigg\}.
\end{align*}
This captures strictly positive (i.e. $\fcn_{j}>0$)
CPDs for the $j$th node that are Lipschitz continuous. 

The parameter spaces $\fcns_{j}(\gr)$ will be used to model CPDs, i.e. $\trued(x_{j}\given \pa(j))=\tf_{j}(x_{j},\pa(j))\in\fcns_{j}(\gr)$.
A DAG $\gr$ defines $d$ such ``local'' function classes 
and a global function class by
\begin{align}
\label{eq:def:fcns}
\fcns(\gr)
:= \fcns_{1}(\gr)\times\cdots\times\fcns_{d}(\gr).
\end{align}
Thus, more formally, given $\fcn\in\fcns(\gr)$, we have $\fcn_{j}\in\fcns_{j}(\gr)$. 
These function classes induce a family of joint density functions as follows. 
The parameter space is taken to be $\fcns(\gr)$ and given $\fcn\in\fcns(\gr)$ (i.e. $d$ functions $\fcn_{j}\in\fcns_{j}(\gr)$), we define a formal function $\jointdf$ as in \eqref{eq:joint:model}. This setup ensures that the resulting $\jointdf$ is always a well-defined joint density.
Define
\begin{align}
\dens
= \dens_{1,\lip}
:=\bigcup_{\gr\in\DAG}\{\jdf: f\in\fcns(\gr)\},
\end{align}
i.e. the collection of all joint densities induced by conditional probabilities in some $\fcns(\gr)$. 
It is easy to check that $\dens_{1,\lip}$ consists of all strictly positive densities in $\lipclass(\I^{d})$. We record this fact here:
\begin{fact}
\label{fact:jpd}
\begin{enumerate*}[label=(\alph*)]
\item\label{jpd:subset} $\dens_{1,\lip}\subset\lipclass(\I^{d})$;
\item\label{jpd:strpos} $\dens_{1,\lip}$ consists of all strictly positive densities in $\lipclass(\I^{d})$; and 
\item\label{jpd:tocpd} $\jointd\in\model(\gr)\cap\dens_{1,\lip}$ implies $\jointd=\jointdf$ for some $\fcn\in\fcns(\gr)$.
\end{enumerate*}
\end{fact}

The preceding discussion can be summarized with the following assumption:
\begin{enumerate}[label=(A\arabic*)]
\item\label{cpd:cond:holder} Each conditional density $\tf_{j}$ is strictly positive and Lipschitz continuous, i.e. $\tf\in\fcns(\tg)$. 
\end{enumerate}

\begin{changemargin}{1cm}{0cm}\vspace{-1.5em} 
\begin{remark}
\label{rem:assm}
These assumptions are made for two primary reasons: First, and foremost, \emph{some} assumptions are needed to make the problem solvable even in principle.
Second, as mentioned above, these assumptions simplify the presentation.
In particular, the use of smoothness assumptions is standard in the literature on nonparametric statistics \citep[see e.g.][]{tsybakov2009introduction,gine2016mathematical,ghosal2017fundamentals}.
It is straightforward to extend our results to more general settings with H\"older continuity, local smoothness, 
and non-compact domains.
\end{remark}
\end{changemargin}

Next, we define metrics on these spaces as follows: The space of joint densities $\dens$ is a metric space under the Hellinger metric 
\begin{align*}
\dH^{2}(\jointd,q)
:= \int_{\I^{d}} \big\{\jointd^{1/2}(x) - q^{1/2}(x)\big\}^{2}\dif{x},
\end{align*}
i.e. the $L^{2}$-distance between the square root densities.
A natural metric on $\fcns_{j}(\gr)$ (i.e. between CPDs) is the integrated Hellinger metric defined by
\begin{align}
\label{eq:hell:int}
\fcnm^{2}(\fcn_{j}, \fcng_{j})
= \int_{\I^{\sp_{j}+1}} \big\{\fcn_{j}^{1/2}(x_{j}\given x_{S}) - \fcng_{j}^{1/2}(x_{j}\given x_{S})\big\}^{2} \truedf(x_{S})\dif{x_{S}}\dif{x_{j}},
\quad\WHERE\quad
S:=\pa_{\gr}(j).
\end{align}
Unless otherwise specified, this is the metric we use.
Finally, we endow $\fcns(\gr)$ with the product metric induced by the $\fcns_{j}(\gr)$, i.e. $\fcnm^{2}(\fcn, \fcng)=\sum_{\ell}\fcnm^{2}(\fcn_{\ell}, \fcng_{\ell})$. We abuse notation by denoting this product metric by $\fcnm$ as well, so that $(\fcns(\gr),\fcnm)$ is a metric space for every $\gr\in\dags$. 

\begin{changemargin}{1cm}{0cm}\vspace{-1.5em} 
\begin{remark}
\label{rem:completion}
We will, from time to time, want to measure the distance between functions with different arguments, e.g. if $\fcn_{j}\in\fcn_{j}(\gr)$ and $\fcng_{j}\in\fcn_{j}(\gr_{kj})$. Strictly speaking, in this case $\fcnm(\fcn_{j},\fcng_{j})$ is not defined, however, we will interpret this as follows in the sequel:
For any $\fcn_{j}\in\fcns_{j}(\gr)$, we define its ``completion'' $\cfcn_{j}:\R^{d}\to\R$ to be the formal $d$-variate function that is equal to $\fcn_{j}$ on $(x,\pa(j))$ and independent of the variables outside $(x_{j},\pa(j))$. Then $\fcnm(\fcn_{j},\fcng_{j}):=\fcnm(\cfcn_{j},\cfcng_{j})$.
This convention helps to avoid some notational gymnastics in having functions with different arguments, and will be used implicitly without further mention in the sequel.
\end{remark}
\end{changemargin}

\subsection{Priors and posteriors}
\label{sec:bg:prior}

To each DAG $\gr$ we will assign a (nonparametric) prior $\pr(\fcn\given\gr)$ over $\fcns(\gr)$ as well as a model prior $\pr(\gr)$ over the space of DAGs.
Moreover, the factorization \eqref{eq:joint:model} provides a model for $P(X\given\fcn)$.
Together, they specify the following Bayesian model:
\begin{align}
\label{eq:bayes:model}
P(X,f,\gr)
= P(X\given f)\pr(f\given\gr)\pr(\gr)
\implies
P(X) = \sum_{\gr\in\DAG}\int_{\fcns(\gr)} P(X\given f)\pr(f\given\gr)\pr(\gr)\dif{f}.
\end{align}
Of course, we are implicitly assuming here necessary measurability assumptions that make this expression well defined.

Unlike classical Bayesian nonparametric models which can be very general in principle, we require specialized structural priors that factorize over neighbourhoods. This 
is key to enabling greedy search and the important property of \emph{decomposability} 
(Definition~\ref{defn:decomp}). 
We encode this in the following definition:
\begin{defn}
\label{defn:strprior} 
We call $\pr(\fcn\given\gr)$ and $\pr(\gr)$ \emph{structural priors} if the following conditions hold for each $\gr\in\dags$:
\begin{align}
\label{eq:cond:prior:1}
\pr(\fcn\given\gr)
&= \pr_{1}(f\given\pa_{\gr}(1))\cdots\pr_{d}(f\given\pa_{\gr}(d)), \\
\label{eq:cond:prior:2}
\pr(\gr)
&= \pr_{1}(\pa_{\gr}(1))\cdots\pr_{d}(\pa_{\gr}(d)),
\end{align}
and
\begin{align}
\label{eq:cond:prior:3}
\pa_{\gr}(j)=\pa_{\grH}(k)
\implies
\left\{
\begin{aligned}
\pr_{j}(\fcn\given\pa_{\gr}(j))
&= \pr_{k}(\fcn\given\pa_{\grH}(k)), \\
\pr_{j}(\pa_{\gr}(j))
&= \pr_{k}(\pa_{\grH}(k)).
\end{aligned}
\right.
\end{align}
\end{defn}
\noindent
Under any choice of structural priors, we have the following useful expression given i.i.d. data $X=(X^{(1)},\ldots,X^{(n)})$:
\begin{align}
P(X\given\gr)
= \int_{\fcns(\gr)} P(X\given\fcn)\pr(\fcn\given\gr)\dif{\fcn}
= \int_{\fcns(\gr)} \prod_{i=1}^{n}\prod_{\ell=1}^{d}\fcn_{\ell}(X_{\ell}^{(i)}\given X_{\pa_{\gr}(\ell)}^{(i)})\pr_{\ell}(\fcn\given\pa_{\gr}(\ell))\dif{\fcn}.
\end{align}

\section{Greedy equivalence search}
\label{sec:ges}

Since the BIC score is no longer well-defined in our nonparametric model, we will replace BIC with an abstract test that compares two DAG models. The main purpose of this section is to show that testing instead of scoring still leads to a correct algorithm.
We defer the construction of such a test to the next section.

\subsection{Original GES}
\label{sec:ogges}

We begin by reviewing the classical GES algorithm that relies on the BIC score to score candidate DAG models. 
GES operates by adding and removing edges according to a scoring criterion---usually BIC---and searches over Markov equivalence classes (MECs), represented by CPDAGs, instead of DAGs. This local search space is defined by the neighbourhoods $\nb_{\pm}(\gr)$ given in (\ref{eq:def:nbplus}-\ref{eq:def:nbminus}).
To provide context for our modifications, and to provide a concrete example, we review the basic steps of GES on a simplified example that searches over DAGs, 
bearing in mind that a more efficient implementation would search over MECs instead of DAGs.

We will denote the sequence of GES iterates by $\gr^{(t)}$.
GES starts with the null model, defined by the empty DAG with $\edg=\emptyset$, corresponding to all possible marginal and conditional independence relationships. That is, $\gr^{(0)}=\nullgr$, where $\nullgr=[\emptyset\given\emptyset\given\cdots\given\emptyset]$ is the empty DAG. Then, for each $t=0,1,2,\ldots,$ we consider all DAGs $\gr\in\nb_{+}(\gr^{(t)})$ and compute the BIC score of each model in this neighbourhood. More precisely, each $\gr\in\nb_{+}(\gr^{(t)})$ defines a model $\model(\gr)$, with corresponding model evidence $P(X\given \model(\gr))$, and assuming the model is sufficiently regular, the evidence can be approximated with the BIC. GES picks the DAG that maximizes this score:
\begin{align}
\label{eq:ges:update}
\gr^{(t+1)}
= \argmax_{\gr\in\nb_{+}(\gr^{(t)})} P(X\given \model(\gr))
\approx \argmax_{\gr\in\nb_{+}(\gr^{(t)})} \BIC(\model(\gr)).
\end{align}
This continues until either the model is saturated (i.e. all edges are included), or there is no DAG in the neighbourhood that increases the BIC score. This is called the \emph{forward phase} of GES.
The second \emph{backward phase} operates analogously, except instead of searching over $\nb_{+}(\gr^{(t)})$ (i.e. larger, more complex models), we search over $\nb_{-}(\gr^{(t)})$ (i.e. smaller, simpler models). Once again, this continues until there is no DAG in the neighbourhood that increases the BIC score. 

Let $\revidx$ denote the last iterate of the forward phase, so e.g. the backward phase is initialized at $\gr^{(\revidx)}$ and $\gr^{(\revidx+1)}$ is the first iterate of the backward phase. The total number of iterations computed by GES will be denoted by $\lastidx$. Thus, we have the following notational conventions:
\begin{itemize}
\item Forward phase: $(\gr^{(0)},\gr^{(1)},\ldots,\gr^{(\revidx)})$
\item Backward phase: $(\gr^{(\revidx)},\gr^{(\revidx+1)},\ldots,\gr^{(\lastidx)})$
\item Full GES search path: $(\gr^{(0)},\ldots,\gr^{(\revidx)},\ldots,\gr^{(\lastidx)})$
\end{itemize}

The main result of \citet{chickering2002ges} is to show that this greedy search procedure is consistent for curved exponential families (e.g. Gaussian and multinomial models). More generally, GES is consistent as long as the BIC score is a reasonable approximation to the model evidence, i.e. as long as the model is sufficiently regular.
Our approach is different: Instead of scoring DAG models, we use pairwise model comparisons, i.e. a test.
This simplifies model comparisons and obviates the need for regularity conditions to justify BIC-style approximations, thereby allowing us to significantly relax the assumptions imposed on the joint density $\trued$. 
More precisely, we use the posterior odds ratio, and re-frame the original consistency proof in terms of the posterior odds. 
The remainder of this section focuses on developing the machinery needed to replace scores with abstract tests.

\subsection{Tests}
\label{sec:tests}

In order to simplify the notation in the sequel, we will first describe the population version of GES: Instead of receiving data $X=(X^{(1)},\ldots,X^{(n)})$ as input, we consider the infinite-sample limit where $\true$ is given as input, and the goal is to output the Markov equivalence class of $\truegr$. Then in Section~\ref{sec:bayes}, we will extend this to a finite-sample version and prove its consistency (Theorem~\ref{thm:rgc}). This is similar to the approach in related work \citep[e.g.][]{kalisch2007,nandy2018}, and hopefully helps to separate population from sample issues.

Let $\test$ be an abstract test that decides which model, $\model$ vs. $\model'$ is a better ``fit'' for an arbitrary distribution $\dist$. We do not assume $\dist\in\model$ or $\dist\in\model'$ and hence allow for misspecification.
In the context of GES, 
we are interested in comparing the current iterate $\gr$ to a new neighbour $\gr'\in\nb_{\pm}(\gr')$.
In order to distinguish population vs. sample aspects, we will consider two kinds of tests: \emph{Abstract} (or population-level) tests that take $\dist$ as input, i.e. $\test(\dist;\gr,\gr')$, and \emph{finite-sample} tests that take the data $X\sim\distn$ as input, i.e. $\test(X;\gr,\gr')$. For clarity, we begin with the population case and then treat the finite-sample case at the end of this section.

More formally, $\test:\dens\times\dags\times\dags\to\{0,1\}$ is a test function that
\emph{compares} two models $\model(\gr)$ and $\model(\gr')$, i.e. $\gr$ is preferred over $\gr'$ if $\test$ selects $\gr$ over $\gr'$, indicated by $\test(P;\gr,\gr')=1$. Otherwise, $\test(P;\gr,\gr')=0$ if $\gr'$ is preferred. This terminology will be used repeatedly in the sequel, so we codify this formally here:
\begin{defn}
\label{defn:preferred}
A DAG $\gr$ is \emph{preferred} (according to $\test$) over another DAG $\gr'$ if $\test(P;\gr,\gr')=1$.
\end{defn}

\citet{chickering2002ges} defined three conditions on a scoring function: 1) Decomposability, 2) (Global) score consistency, and 3) Local score consistency. As long as the scoring function satisfies these three conditions, the original score-based version of GES is guaranteed to consistently recover the underlying Markov equivalence class.
To extend these definitions to tests $\test$ as defined above, we will use
the following natural translation of these properties from scores to tests:

\begin{defn}\label{defn:decomp}
A test $\test(P;\gr,\gr')$ is called \emph{decomposable} if for any $\dist$ and any $\gr,\grH\in\DAG$ with $\pa_{\grH_{-kj}}(j)=\pa_{\gr_{-kj}}(j)$ (and hence also $\pa_{\grH_{kj}}(j)=\pa_{\gr_{kj}}(j)$), it follows that 
\begin{align}
\label{eq:defn:decomp}
\test(P;\gr_{kj},\gr_{-kj})
= \test(P;\grH_{kj},\grH_{-kj}).
\end{align}
\end{defn}
\noindent
In other words, a test is decomposable if checking whether or not to include or remove an edge $k\to j$ depends only on the parents of $j$.

\begin{defn}
\label{defn:global_consistency}
A test $\test$ is called \emph{globally consistent} if it satisfies the following conditions:
\begin{enumerate}[label=(G\arabic*)]
\item\label{defn:global_consistency:1} If $P\in\model(\gr)$ and $P\notin\model(\grH)$, then $\test(P;\gr,\grH)=1$.
\item\label{defn:global_consistency:2} If $P\in\model(\gr)\cap\model(\grH)$, and $|\edg(\gr)| < |\edg(\grH)|$, then $\test(P;\gr,\grH)=1$.
\end{enumerate}
We call $\test$ \emph{1-globally consistent} if these conditions hold whenever $\gr$ and $\grH$ are neighbours, i.e. $\grH\in\nb_{\pm}(\gr)$.
\end{defn}

\begin{defn}
\label{defn:local_consistency}
Let $\gr$ be any DAG such that (a) $\gr=\gr_{-kj}$, i.e. $k\to j$ is not in $\gr$, and (b) $\gr_{kj}$ is also a DAG. A test $\test$ is called \emph{locally consistent} with respect to $(P,\gr)$ if it satisfies the following conditions:
\begin{enumerate}[label=(L\arabic*)]
\item\label{defn:local_consistency:1} If $k\not\indepP j\given \pa_{\gr}(j)$ then $$\test(P;\gr_{kj},\gr)=1\iff\test(P;\gr,\gr_{kj})=0.$$
In other words, given the parents in $\gr$, if $X_{k}$ and $X_{j}$ are still dependent, then the test prefers DAGs with the edge $k\to j$ added.
\item\label{defn:local_consistency:2} If $k\indepP j\given \pa_{\gr}(j)$ then $$\test(P;\gr_{kj},\gr)=0\iff\test(P;\gr,\gr_{kj})=1.$$
In other words, given the parents in $\gr$, if $X_{k}$ and $X_{j}$ are independent, then the test prefers DAGs with the edge $k\to j$ removed.
\end{enumerate}
\end{defn}

The following lemma shows that only two of these conditions are truly necessary:
\begin{lemma}
\label{lem:global2local}
If a test $\test$ is 1-globally consistent and decomposable, then it is locally consistent. 
\end{lemma}

So far, we have assumed that we are given the distribution $\dist$ as input.
A finite-sample test $\test(X;\gr,\gr')$ is entirely analogous to the abstract tests defined above, except it uses only the data $X$ instead of $\dist$. Definitions~\ref{defn:preferred}-\ref{defn:local_consistency} carry over for finite-sample tests in the obvious way, with all equalities interpreted as convergence in probability as $n\to\infty$. Thus, for example, Definition~\ref{defn:preferred} is interpreted as: A DAG $\gr$ is \emph{(asymptotically) preferred} (according to $\test$) over another DAG $\gr'$ if 
\begin{align*}
\test(X;\gr,\gr')\overset{\dist}{\to} 1
\quad\AS\,
n\to\infty,
\quad\WHERE\,
X\sim \dist^{n}.
\end{align*}
In our theorem statements, we use the modifier ``asymptotically'' to emphasize when $\test$ is to be interpreted at the sample level. In most cases, it will be clear from context already.

\begin{changemargin}{1cm}{0cm}\vspace{-1.5em} 
\begin{remark}
\label{rem:testerr}
We have 
\begin{align*}
\test(X;\gr,\gr')\overset{\dist}{\to} 1
\iff
\E[1-\test(X;\gr,\gr')]\to 0,
\end{align*}
where $\E[1-\test(X;\gr,\gr')]$ is the usual testing error. Thus, both global and local consistency can be interpreted as requirements on the testing error of $\test$. The requirement \ref{defn:global_consistency:1} in global consistency is equivalent to $\test$ being \emph{consistent} in testing $\model(\gr)$ vs $\model(\grH)$. Local consistency is more subtle, since we do not require $P\in\model(\gr)$ or $P\in\model(\gr_{kj})$; i.e. the test is misspecified. Formally, local consistency requires consistency at the level of a single CI relation, even though the global model may itself be misspecified. For more on the connection with CI testing, which is subtle, see Remark~\ref{rem:citest}.
\end{remark}
\end{changemargin}

\begin{changemargin}{1cm}{0cm}\vspace{-1.5em} 
\begin{remark}
\label{rem:citest}
It is tempting to presume that comparing $\model(\gr)$ vs $\model(\gr_{kj})$ is equivalent to testing the single CI relation $k\indepP j\given \pa_{\gr}(j)$, however, this is false in general. For example, suppose that in the first iteration we add the edge $1\to2$. Then the model given by the $v$-structure $1\to2\leftarrow3$ will be in the neighbourhood of this state, and these two models encode multiple CI relations: The single-edge model encodes (at least) the CI relations $\{1\indep 3, 1\indep 3\given 2, 2\indep 3, 2\indep3\given 1\}$, whereas the $v$-structure only implies $\{1\indep 3\}$. In other words, testing $\model(\gr)$ vs $\model(\gr_{kj})$ involves simultaneously testing three CI relations given by $\{1\indep 3\given 2, 2\indep 3, 2\indep3\given 1\}$.
\end{remark}
\end{changemargin}

\begin{program}[t]
\begin{minipage}[t]{0.48\textwidth}
\begin{algorithm}[H]
\caption{Forward phase}
\DontPrintSemicolon
\KwIn{Distribution $\true$}
Initialize $t \leftarrow 0$, $\gr^{(t)}\leftarrow\gr_{\emptyset}$\;
\Repeat{$\gr^{(t+1)}=\gr^{(t)}$}{
    $\gr^{(t+1)}\leftarrow\gr^{(t)}$\;
    \For{$\gr\in\nb_{+}(\gr^{(t)})$}{
        \If{$\test(\true;\gr,\gr^{(t)})=1$}{$\gr^{(t+1)}\leftarrow\gr$; \Break}
}
    $t \leftarrow t+1$\;
}
\textbf{set} $\revidx\leftarrow t$\;
\KwOut{DAG $\gr^{(\revidx)}$}
\end{algorithm}
\end{minipage}
\begin{minipage}[t]{0.48\textwidth}
\begin{algorithm}[H]
\caption{Backward phase}
\DontPrintSemicolon
\KwIn{Distribution $\true$}
Initialize $t \leftarrow \revidx$, $\gr^{(t)}\leftarrow\gr^{(\revidx)}$\;
\Repeat{$\gr^{(t+1)}=\gr^{(t)}$}{
    $\gr^{(t+1)}\leftarrow\gr^{(t)}$\;
    \For{$\gr\in\nb_{-}(\gr^{(t)})$}{
        \If{$\test(\true;\gr,\gr^{(t)})=1$}{$\gr^{(t+1)}\leftarrow\gr$; \Break}
}
    $t \leftarrow t+1$\;
}
\textbf{set} $r\leftarrow t$\;
\KwOut{DAG $\gr^{(\lastidx)}$}
\end{algorithm}
\end{minipage}
\caption{Pseudocode for the modified GES algorithm based on testing instead of scoring. The forward phase (left) and backward phase (right) are each run once.}\label{alg:ges}
\end{program}

\subsection{Modified GES}
\label{sec:modges}

Since we can no longer score candidate DAG models with BIC, the original update scheme for GES in \eqref{eq:ges:update} does not apply. Fortunately, there is a simple modification that turns out to suffice: Instead of searching for the highest scoring DAG, during the forward phase we simply accept \emph{any} neighbour $\gr\in\nb_{+}(\gr^{(t)})$ that is preferred over $\gr^{(t)}$ (cf. Definition~\ref{defn:preferred}).
During the backward phase, we accept any neighbour $\gr\in\nb_{-}(\gr^{(t)})$ that is preferred. 

Pseudocode of this ``modified GES'' is provided in Algorithm~\ref{alg:ges}, which we will often refer to as simply ``GES'' in the sequel.
The following result proves the correctness of this algorithm at the population level, and is essentially a formal extension of \citet{chickering2002ges} to tests instead of scores.
\begin{prop}
\label{prop:ges:correct}
Assume that $\test$ is both locally and 1-globally consistent. Then the output of Algorithm~\ref{alg:ges} is a perfect map of $\true$, i.e. $\ind(\gr^{(\lastidx)}) = \ind(\true)$.
\end{prop}
\noindent

To instantiate this result for nonparametric DAG models, by Lemma~\ref{lem:global2local}, it suffices to construct a test that is both decomposable and 1-globally consistent. For regular parametric models, this is straightforward. For nonparametric models, this is nontrivial: 
While there is a robust literature for the simpler problem of testing a single CI relation, developing a fully nonparametric test for comparing two nonparametric graphical models is an open problem.

\section{Bayesian model selection for nonparametric graphical models}
\label{sec:bayes}

Proposition~\ref{prop:ges:correct} is an abstract result that outlines a set of sufficient conditions that ensure the correctness of GES (Algorithm~\ref{alg:ges}) given $\true$ as input. In practice, we are given i.i.d. samples $X^{(i)}\sim\true$, and we wish to prove the consistency of GES when $\test$ is replaced by a finite-sample test.

Our approach will be to define an appropriate Bayesian nonparametric model, and use the resulting posterior to construct a consistent test that is both decomposable and 1-globally consistent. While the consistency of Bayesian tests in general nonparametric settings is known \citep{lember2007universal,ghosal2008nonparametric,mcvinish2009bayesian,yang2017bayesian}, graphical models present additional technical challenges. 
In particular, instead of working with the joint density directly as in previous work, we work with the Markov factorization directly through the $d$ conditional densities, parametrized by $\fcn=(\fcn_{1},\ldots,\fcn_{d})$. This is crucial to obtaining ``local'' guarantees involving single edge comparisons. 
An additional technical complication is obtaining an explicit comparison of the complexity between models. 
We outline these steps in the present section.

\subsection{Bayesian tests}
\label{sec:decomp}

Recalling the setup in Section~\ref{sec:bg}, for each DAG $\gr$ we have a corresponding function space $\fcns(\gr)$ that maps into the family of joint densities $\dens$. Given a prior $\pr(\fcn\given\gr)$ over $\fcns(\gr)$ as well as a model prior $\pr(\gr)$ over $\DAG$, \eqref{eq:bayes:model} defines 
the (parameter) posterior $P(\fcn\given X)$, the model posterior $P(\gr\given X)$, and the model evidence $P(X\given\gr)$.

Given two competing models $\gr$ and $\grH$,
define the posterior odds ratio as follows:
\begin{align}
\label{eq:defn:odds}
\odds(\gr,\grH)
= \frac{P(\gr\given X)}{P(\grH\given X)},
\quad\WHERE\quad
P(\gr\given X)
\propto \int_{\fcns(\gr)} P(X\given\fcn)\pr(\fcn\given\gr)\pr(\gr)\dif{\fcn}.
\end{align}
This is related to the well-known Bayes factor $\BF(\gr,\grH)$ by 
\begin{align*}
\odds(\gr,\grH)
= \frac{\pr(\gr)}{\pr(\grH)}\cdot\BF(\gr,\grH),
\quad\WHERE\quad
\BF(\gr,\grH)
= \frac{P(X\given \gr)}{P(X\given \grH)}
= \frac{\int_{\fcns(\gr)} P(X\given \fcn)\pr(\fcn\given\gr)\dif{\fcn}}{\int_{\fcns(\grH)} P(X\given \fcn)\pr(\fcn\given \grH)\dif{\fcn}}.
\end{align*}
This defines a test $\test$ by simply thresholding the posterior odds at some $\level>0$:
\begin{align}
\label{eq:test:odds}
\test_{\level}(X;\gr,\grH)
= \test(\gr,\grH)
:= 1(\odds(\gr,\grH) > \level).
\end{align}
We will often suppress the dependence of $\test$ on $X$ and $\level$ for brevity.

By operating on the space of CPDs and using structural priors (Definition~\ref{defn:strprior}), 
we obtain simultaneous local (i.e. conditional) and global (i.e. joint) control over the parameters in the model. As a result, we can show that the thresholding test $\test$ defined in \eqref{eq:test:odds} is decomposable:
\begin{lemma}
\label{lem:bf:decomp}
Under any choice of structural priors (\ref{eq:cond:prior:1}-\ref{eq:cond:prior:3}), the test $\test_{\level}$ defined in \eqref{eq:test:odds} satisfies $\odds(\gr_{kj},\gr_{-kj})=\odds(\grH_{kj},\grH_{-kj})$ for any $\gr,\grH\in\DAG$ with $\pa_{\grH_{-kj}}(j)=\pa_{\gr_{-kj}}(j)$. In particular, it is decomposable for any $\level>0$.
\end{lemma}
\noindent
Unlike most of our claims involving finite-sample tests, this is not an asymptotic statement: This holds for all finite $n$ and any data $X$.

\subsection{Prior assumptions}
\label{sec:priors}

The next step is to establish the (global) consistency of $\test$ under structural priors $\pr(\fcn\given\gr)$ and $\pr(\gr)$. 
For this,
we choose individual mixture priors for each component $j\in\ver$ separately and then combine them.
We use a Dirichlet process mixture of gaussian prior over each $\fcns_{\ell}(\gr)$ with an inverse Gamma prior over the scale parameter, as in \citet{shen2013adaptive,norets2017adaptive}.
Given these priors $\pr_{\ell}(\fcn_{\ell}\given\pa_{\gr}(\ell))$ over $\fcns_{\ell}(\gr)$, we define a prior over $\fcns(\gr)$ by
\begin{align}
\label{eq:prior:fcngr}
\pr(\fcn\given\gr)
= \prod_{\ell=1}^{d} \pr_{\ell}(\fcn_{\ell}\given\pa_{\gr}(\ell)).
\end{align}
Next, define a prior over models (i.e. over $\dags$) by choosing constants $\prconstj_{\gr}>0$ and $\prconst>0$, and defining
\begin{align}
\label{eq:prior:gr}
\pr(\gr)
\propto \prconstj_{\gr}e^{-\prconst n\eps_{n}^{2}(\gr)},
\end{align}
where
$\me_{n}(\gr)>0$ is the unique positive solution to the following polynomial equation in $\eps$: 
\begin{align}
\label{eq:me:poly}
\sum_{\ell=1}^{d}\Big(\frac{d}{\eps}\Big)^{(\sp_{\ell}(\gr)+1)/\reg}
= n\eps^{2}.
\end{align}
Recall that to each DAG $\gr$ we have its sparsity sequence $\sp(\gr)=(\sp_{1}(\gr),\ldots,\sp_{d}(\gr))$.

\begin{lemma}
\label{lem:cond:prior}
The priors $\pr(\fcn\given\gr)$ and $\pr(\gr)$ defined in (\ref{eq:prior:fcngr}-\ref{eq:prior:gr}) are structural priors.
\end{lemma}

The polynomial equation \eqref{eq:me:poly} is Le Cam's equation, and arises in the analysis of nonparametric Bayesian models \citep{ghosal2017fundamentals}. The solutions to this equation are a measure of regularity of the underlying graphical model, with more complex graphs (i.e. more edges) yielding larger $\me_{n}(\gr)$. 
Although we do not have closed-form expressions for the solutions $\me_{n}(\gr)$, what matters in the proofs is that these numbers appropriately discriminate models of different complexities. 
In practical implementations (e.g. in constructing the prior \eqref{eq:prior:gr}), these numbers can be numerically approximated to arbitrary accuracy. For now, we collect some useful properties of these numbers:
\begin{fact}
The sequences $\me_{n}(\gr)$ have the following properties:
\begin{enumerate}
\item $\me_{n}(\gr)\to 0$ as $n\to\infty$;
\item $n\me_{n}^{2}(\gr)\to \infty$ as $n\to\infty$;
\item $\me_{n}(\gr)<\me_{n}(\gr')$ for any $\gr'\in\nb_{+}(\gr)$.
\end{enumerate}
\end{fact}
\noindent

\subsection{Model separation}
\label{sec:gap}

Global consistency, as stated in Definition~\ref{defn:global_consistency}, allows that $\dist\in\model(\gr)$ is arbitrarily close to $\model(\grH)$.
Of course, we do not expect to be able to distinguish $\gr$ from $\grH$ on the basis of samples from $\dist$ if these two models are too close to one another. In order to ensure these models can be distinguished with data, there must be some separation between these models.
In the context of testing, this is well-known \citep[see e.g.][]{balakrishnan2018hypothesis,neykov2020minimax}. Similar conditions appear as well in the literature on structure learning \citep[e.g.][]{loh2014causal}, and are closely related to beta-min conditions in variable selection \citep{yang2017bayesian,liu2021variable}.

Motivated by the need for separation between models, we propose to quantify separation via the metric $\fcnm$ on $\fcns(\gr)$. Given any distribution $\dist=\distf$ and any DAG $\gr$, we can ask how far $\fcn$ is from the corresponding model $\fcns(\gr)$, as measured by the metric $\fcnm$ (cf. \eqref{eq:hell:int}, see also Remark~\ref{rem:completion}):
\begin{align}
\label{eq:defn:signal}
\fcnm(\fcn,\gr)
= \fcnm(\fcn,\fcns(\gr))
:= \inf_{\fcng\in\fcns(\gr)} \fcnm(\fcn,\fcng).
\end{align}
\noindent
If $f\in\fcns(\gr)$ then $\fcnm(\fcn,\gr)=0$, and $\fcnm(\fcn,\gr)>0$ implies $f\notin\fcns(\gr)$. Thus, $\fcnm(\fcn,\gr)$ measures how close $\fcn$ is to $\fcns(\gr)$. Phrased differently, $\fcnm(\fcn,\gr)$(implicitly) measures how close $\jointdf$ is to satisfying the CI relations implied by $\gr$.

More precisely, \eqref{eq:defn:signal} also captures an intuitive measure of dependence. To see this, fix a DAG $\gr$ and choose a pair of nodes $X_{k}$ and $X_{j}$ such that $(k,j)\notin\edg(\gr)$, but $k\not\indepx{\dist} j\given \pa_{\gr}(j)$, so that $\fcn\notin\fcns(\gr)$. To simplify calculations, let's assume that $\fcn\in\fcns(\gr_{kj})$, so that $\fcnm(\fcn,\gr)>0$ quantifies how far the model $\fcns(\gr)$ is from the ``correct'' model $\fcns(\gr_{kj})$. Using $\fcnm(\fcn,\gr_{kj})=0$ and also $\fcnm(\fcn_{\ell},\fcns_{\ell}(\gr))=0$ for $\ell\ne j$, we see that the difference
\begin{align*}
\fcnm(\fcn,\gr) - \fcnm(\fcn,\gr_{kj}) 
= \fcnm(\fcn,\gr) 
= \sum_{\ell=1}^{d}\fcnm(\fcn_{\ell},\fcns_{\ell}(\gr))
= \fcnm(\fcn_{j},\fcns_{j}(\gr))
> 0
\end{align*}
quantifies how much we lose by ``ignoring'' the edge $k\to j$, despite knowing that these variables are conditionally dependent. In other words, it implicitly measures the strength of the conditional dependence between $k$ and $j$, given $\pa_{\gr}(j)$. Alternatively, this can be interpreted via the lens of local consistency: The (correct) presence of $k\to j$ in $\gr_{kj}$ indicates that $k\not\indepx{\dist} j\given \pa_{\gr}(j)$ and hence \ref{defn:local_consistency:1} implies that $\gr_{kj}$ is preferred over $\gr$. Then $\fcnm(\fcn_{j},\fcns_{j}(\gr))$ measures the signal needed to satisfy local consistency, i.e. to detect this edge.

With these examples in mind, in order to distinguish $\gr$ and $\gr'$, we assume a lower bound on $\fcnm(\tf,\gr)$ on the order $\me_{n}(\gr')$.
More precisely:
\begin{enumerate}[label=(A\arabic*),start=2]
\item\label{cond:gap} For any $\gr\in\dags$ and $\gr'\in\nb_{+}(\gr)$ such that $X_{k}\not\indepx{\true} X_{j}\given \pa_{\gr}(j)$, we have $\fcnm(\tf,\gr)\ge \omega(\me_{n}(\gr'))$ as $n\to\infty$.
\end{enumerate}
This essentially captures the minimum signal strength needed to distinguish two models: When $\tf\not\in\fcns(\gr)$ (this is captured by the missing edge $(k,j)\notin\edg(\gr)$), the strength of the missing dependence between $k$ and $j$ should be on the order $\me_{n}(\gr_{kj})$. For the backward phase, which involves comparing to neighbours $\gr'\in\nb_{-}(\gr)$, \ref{cond:gap} still applies by swapping the roles of $\gr$ and $\gr'$.

\subsection{Model selection consistency}

Under the model just constructed, we have the following crucial model selection consistency result:

\begin{thm}
\label{thm:rgc}
Assume \ref{cpd:cond:holder}-\ref{cond:gap} along with the priors constructed above. 
Then for any $\level>0$, $\test_{\level}(X;\gr,\gr')$ is asymptotically 1-globally consistent. 
\end{thm}

Theorem~\ref{thm:rgc}, combined with Lemma~\ref{lem:bf:decomp}, provides a constructive proof of the existence of a general Bayesian nonparametric test for comparing two nonparametric DAG models that is both decomposable and 1-globally consistent. In particular, by Lemma~\ref{lem:global2local}, it is also locally consistent.

\begin{cor}
Under the assumptions of Theorem~\ref{thm:rgc}, $\test_{\level}(X;\gr,\gr')$ is  (asymptotically) decomposable, locally consistent, and 1-globally consistent. 
\end{cor}

\section{Structure consistency}
\label{sec:struct}

Adopting the general setup from the previous sections,
define the test $\test_{\level}$ as in \eqref{eq:test:odds} using the priors defined in the previous section (cf. \ref{eq:prior:fcngr}-\ref{eq:prior:gr}), and let $\gesout=\gesout(\level)$ denote the output of GES (Algorithm~\ref{alg:ges}) using the data-dependent test $\test_{\level}$ with threshold $\level>0$. 
We have the following nonparametric structure consistency result for GES when applied to data (as opposed to the true distribution $\true$).

\begin{thm}
\label{thm:ges:cons}
Assume \ref{cpd:cond:holder}-\ref{cond:gap}. 
Then for any $\level>0$, $\gesout$ converges to a perfect map of $\true$, i.e.
\begin{align*}
\Pr(\ind(\gesout) = \ind(\true)) \to 1.
\end{align*}
\end{thm}
\noindent
For the case where $\true$ does not admit a perfect map, GES still finds a minimal model.

A few aspects of this result are noteworthy:
\begin{itemize}
\item We do not prove that $\gesout=\tg$ because
$\tg$ is not identifiable from $\true$. The conclusion that $\ind(\gesout) = \ind(\true)$ implies that $\ind(\gesout) = \ind(\tg)$, i.e. $\gesout$ is in the same Markov equivalence class as $\tg$. This is the same guarantee as in previous work, e.g. \citet{chickering2002ges,nandy2018}. 
As pointed out previously, faithfulness is not required and is only assumed to simplify the main result. Without faithfulness, GES returns a minimal model, and when a perfect map exists it must be the unique minimal model.
\item In the forward phase, essentially all tests are misspecified. For example, in the first iteration, except in trivial cases, neither $\gr^{(0)}$ nor any of its neighbours will contain $\true$. In this sense, local consistency is a local guarantee on how tests handle misspecification. See also Remark~\ref{rem:citest} on how these tests differ from conventional CI tests.
\end{itemize}

\noindent
As stated, Theorem~\ref{thm:ges:cons} already appears to be one of the most general consistency results available for greedy structure learning, and indeed it can be generalized further; cf. Remark~\ref{rem:assm}.

\section{Discussion}
\label{sec:disc}

Greedy search is a classical strategy for model selection in combinatorial models, and has been widely applied to subset selection, Markov random fields, and most relevant to the present work, DAG models. In the latter case, GES is a foundational approach to score-based learning. Despite several decades of research, our understanding of GES in nonparametric settings has been limited. We have attempted to fill this gap by developing a theoretical framework for analyzing greedy algorithms for model selection, and applying it to establish the consistency of GES for general nonparametric DAG models.

\bibliography{genbib,personalbib}
\bibliographystyle{unsrtnat-nourl}

\end{document}